\RequirePackage[svgnames]{xcolor}

\documentclass[11pt,letterpaper]{include/mystyle}

\usepackage[all]{hypcap}
\usepackage[svgnames]{xcolor}
\usepackage[comma,authoryear,compress]{natbib}
\hypersetup{
    colorlinks = true,
    citecolor = {YaleBlue},
}

\usepackage{microtype}
\usepackage{graphicx}
\expandafter\def\csname ver@subfig.sty\endcsname{}
\usepackage{booktabs} %
\usepackage{multirow} 
\usepackage{float}
\usepackage{bigstrut}

\usepackage{amsmath}
\usepackage{amssymb}
\usepackage{mathtools}
\usepackage{amsthm}
\usepackage{mathrsfs}
\usepackage{nicefrac}
\usepackage{dsfont}
\usepackage{enumitem}
\usepackage{cleveref}
\usepackage{bxcoloremoji}

\usepackage{float}

\usepackage[utf8]{inputenc} %
\usepackage[T1]{fontenc}    %
\usepackage{url}            %
\usepackage{booktabs}       %
\usepackage{amsfonts}       %
\usepackage{nicefrac}       %
\usepackage{microtype}      %
\usepackage{graphicx}
\usepackage{amssymb}
\usepackage{wrapfig}
\usepackage{lipsum}
\usepackage{enumitem}
\usepackage{stackengine}
\usepackage[font=small,labelfont=bf]{caption}
\usepackage{color}
\usepackage{adjustbox}

\usepackage{rotating}
\usepackage{makecell}

\usepackage{amsmath}

\usepackage[all]{hypcap}

\definecolor{blanchedalmond}{rgb}{1.0, 0.92, 0.8}
\definecolor{carmine}{rgb}{0.59, 0.0, 0.09}
\definecolor{lightblue}{rgb}{0.22,0.45,0.70}%

\renewcommand{\mathbf}{\boldsymbol}

\makeatletter
\def\Ddots{\mathinner{\mkern1mu\raise\p@
\vbox{\kern7\p@\hbox{.}}\mkern2mu
\raise4\p@\hbox{.}\mkern2mu\raise7\p@\hbox{.}\mkern1mu}}
\makeatother

\definecolor{amaranth}{rgb}{0.9, 0.17, 0.31}
\definecolor{antiquebrass}{rgb}{0.8, 0.58, 0.46}
\definecolor{antiquefuchsia}{rgb}{0.57, 0.36, 0.51}
\definecolor{chromeyellow}{rgb}{0.31, 0.47, 0.26}

\tcbuselibrary{most}

\newtcolorbox{AIbox}[2][]{aibox,title=#2,#1}
\definecolor{lightblue}{rgb}{0.22,0.45,0.70}%
\definecolor{Gray}{gray}{0.95}
\definecolor{Cornsilk}{rgb}{1.0, 0.97, 0.86}

\usepackage{microtype}
\usepackage{fontawesome}
\usepackage{hyperref}
\usepackage{url}
\usepackage{booktabs}
\usepackage{graphicx}
\usepackage{lineno}
\usepackage{amsmath}
\usepackage{booktabs}
\usepackage{pifont}

\usepackage{booktabs}
\usepackage{pifont}
\usepackage{xcolor}

\usepackage{tcolorbox}
\usepackage{wrapfig}
\tcbset{
    colback=gray!5,
    colframe=black,
    boxrule=0.5pt,
    arc=2pt,
    left=6pt,
    right=6pt,
    top=6pt,
    bottom=6pt
}

\definecolor{darkblue}{rgb}{0, 0, 0.5}
\hypersetup{colorlinks=true, citecolor=darkblue, linkcolor=darkblue, urlcolor=darkblue, hypertexnames=false}
\usepackage{algorithm}
\usepackage{algpseudocode}
\usepackage{enumitem}

\title{When AI Reviews Train AI Reviewers: Scientific-Judgment Collapse and Mitigation}
\author{Anonymous Authors}

\runningtitle{When AI Reviews Train AI Reviewers: Scientific-Judgment Collapse and Mitigation}

\author{%
  Sy-Tuyen Ho*, Minghui Liu*, Hieu Tran, Furong Huang \\
  University of Maryland, College Park\\
  * Equal contribution\\
}

\correspondingauthor{Furong Huang; Email \href{mailto:furongh@umd.edu}{furongh@umd.edu}}

\begin{document}

\begin{abstract}
LLMs increasingly assist scientific peer review. This creates a feedback loop when generated reviews enter the training data of future LLMs. While recursive training has been studied broadly, its effects on scientific judgments remain poorly understood. We empirically study this effect in a controlled setting by varying synthetic-review exposure while fixing other training conditions. After one recursive step, increasing exposure from $0\%$ to $100\%$ makes recommendation ratings more concentrated, independently generated reviews of the same paper about $11\%$ more similar, and reviews across different papers about $5\%$ less diverse. A further step of recursive training amplifies homogenization when synthetic supervision is reintroduced. We characterize this task-specific phenomenon as $\textbf{scientific-judgment collapse}$. We also explore a training-free method using contrastive activation steering with a limited reference reviews. Controlled experiments show a partial recovery in recommendation agreement and judgment diversity across different papers, but not among independently generated reviews of the same paper. To explore this approach beyond controlled experiments, we develop $\textbf{TrustReviewer}$. We train the model on a large-scale curated data and apply activation steering at inference time. On held-out papers, steering increases exact recommendation match from $73.85\%$ to $75.40\%$, compared with $73.10\%$ for OpenReviewer on the same LLM backbone, while maintaining relatively diverse scientific judgments. Our findings quantify judgment homogenization under recursive training and motivate further research on preserving judgment diversity.

\textbf{Project Page} \faHome\textbf{:} \url{https://hosytuyen.github.io/projects/TrustReviewer}

\textbf{Dataset \includegraphics[height=0.4cm]{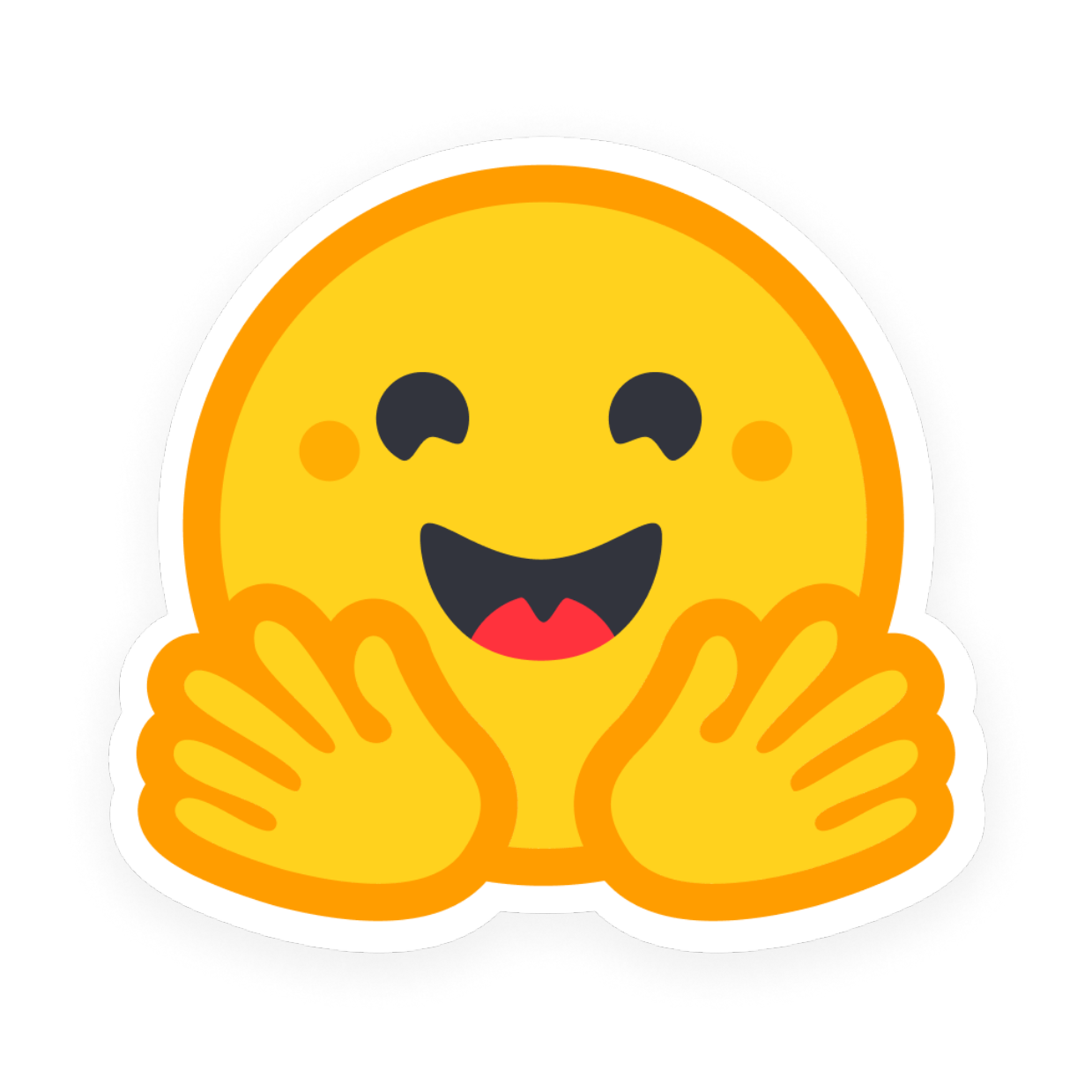} :} \url{https://huggingface.co/collections/hosytuyen/trustreviewer}

\textbf{Code \faGithub:} \url{https://github.com/hosytuyen/TrustReviewer}
\end{abstract}
\maketitle
\section{Introduction}
\label{sec:introduction}

\begin{figure*}[t]
\centering
\includegraphics[width=\linewidth]{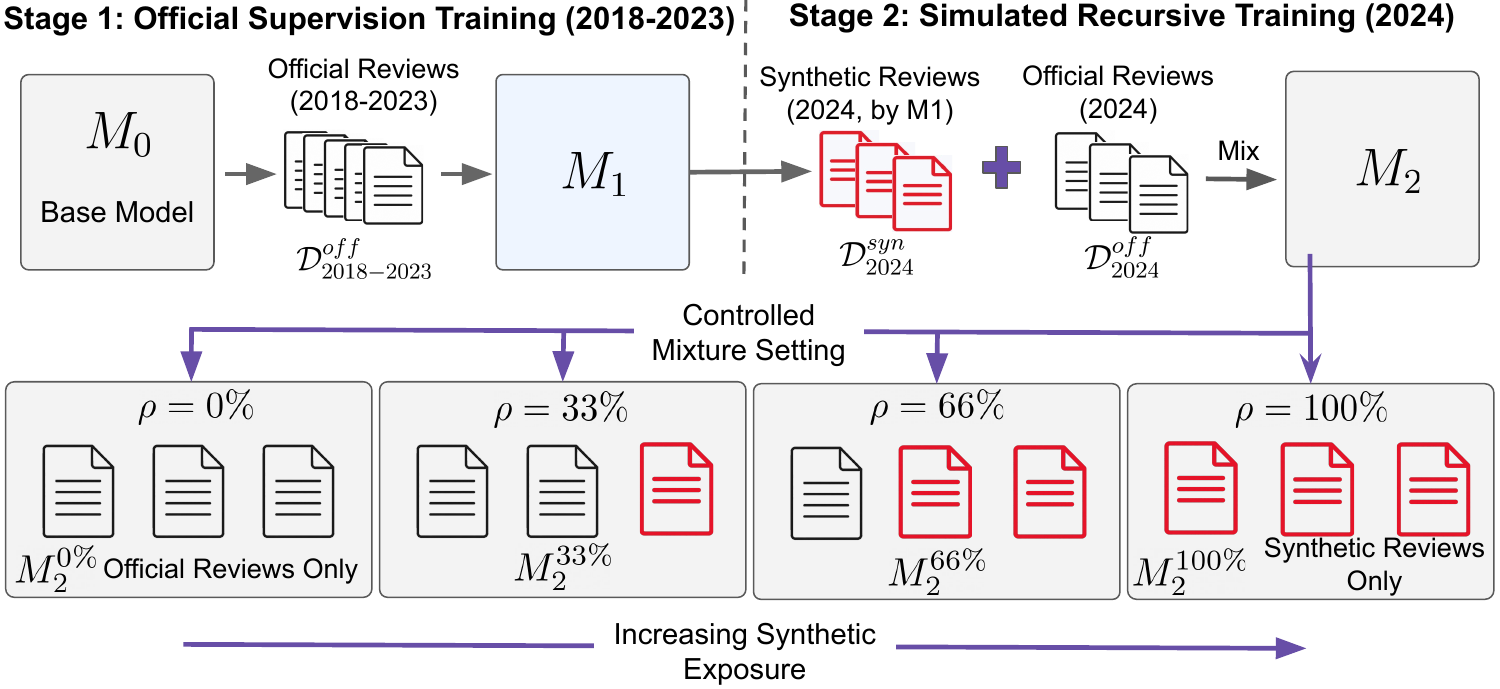}
\caption{\textbf{Controlled framework for recursive AI-reviewer training.}
Starting from the same $M_1$ reviewer, we construct $M_2^{p}$ variants with 0\%, 33\%, 66\%, or 100\% explicitly generated reviews while holding the initialization, review count, filtering, and training configuration fixed.}%
\label{fig:recursive-pipeline}
\end{figure*}

Large language models (LLMs) are increasingly used in scholarly research to summarize papers, suggest critiques, predict recommendation ratings, and help reviewers improve their reports. \citet{liang2024monitoring} provide evidence that substantial LLM-generated modifications are already present in some AI-conference reviews. These uses can reduce reviewer workload but also create a feedback loop. Reviews written or substantially edited by LLMs may become publicly available and subsequently enter the training corpora of future LLMs. This feedback loop creates a form of recursive training. Language-model studies find that increasing the ratio of generated data can make model outputs more concentrated \citep{zhu2024synthesize}. Other studies find that when models are trained on generated data, they can lose low-probability parts of the original data distribution \citep{shumailov2023curse,alemohammad2024self}. These findings raise a concrete question for peer review: \textit{if an LLM is trained on LLM-generated reviews, does it produce a narrower range of scientific judgments and suppress less common judgments?}

We conduct a controlled empirical study to quantify  how this recursive training affects scientific judgments. Starting from a base LLM, i.e. Llama 3.1 8B Instruct \citep{grattafiori2024llama}, we first fine-tune an initial reviewer on the official ICLR reviews from 2018--2023. We then use this reviewer to generate reviews for ICLR~2024 papers and train successor reviewers on mixtures with different ratios of official and generated reviews. We keep the initialization, data processing, and training procedure fixed, so the planned difference across successor models is the source mixture of their training reviews. We further extend the experiment by one more recursive step, comparing successor models with and without prior synthetic exposure at the same synthetic exposure in the current training stage. Because official reviews may contain unobserved AI assistance, our exposure percentages refer only to reviews explicitly generated by the preceding model.

After one recursive step, increasing synthetic exposure from 0\% to 100\% makes recommendation ratings more concentrated. For example, the proportion of rating 6 increases from 33.4\% to 40.0\%, while rating 3 decreases from 20.1\% to 11.2\%. Rating entropy also decreases from 2.31 to 2.18, indicating a narrower distribution of recommendations. Independently generated reviews of the same paper become about 11\% more similar, and reviews across different papers become about 5\% less diverse. In the two-step comparison, prior synthetic exposure amplifies judgment homogenization when synthetic supervision is reintroduced, while the two lineages remain broadly similar in semantic diversity when current-stage synthetic exposure is 0\%. We call this task-specific phenomenon \textbf{scientific-judgment collapse}.

We also explore whether scientific-judgment collapse can be partially mitigated at inference time using limited reference supervision. Specifically, we apply contrastive activation steering to guide the collapsed model toward more diverse judgments. Our controlled experiments show partial recovery in recommendation agreement and semantic diversity across papers, although semantic diversity among independently generated reviews of the same paper decreases. Finally, we extend this steering approach beyond controlled experiments to a practical system, \textbf{TrustReviewer}. We train the model on a large-scale curated data and apply steering at inference time. On held-out evaluation papers, steering increases exact recommendation match from 73.85\% to 75.40\%, compared with 73.10\% for OpenReviewer on the same LLM backbone, while maintaining competitive semantic diversity. In summary, our contributions are:

\begin{enumerate}
    \item \textbf{A controlled study of effect of recursive reviewer training on peer reviews (Secs.~\ref{sec:recursive-study} and \ref{Sec:Scientific-Judgment Collapse under Recursive Training}).}
    We show that synthetic-review exposure concentrates recommendation ratings and reduces semantic diversity. A further recursive step amplifies judgment homogenization when synthetic supervision is reintroduced.

    \item \textbf{An exploratory limited-reference steering approach (Sec.~\ref{sec:inference-time-recovery}).}
    We investigate contrastive activation steering without retraining, finding partial recovery in recommendation agreement and judgment diversity across papers, but not among independently generated reviews of the same paper.

    \item \textbf{TrustReviewer: A practical reviewer system (Sec.~\ref{sec:trustreviewer}).}
    We extend steering approach beyond controlled experiments to a reviewer trained on large-scale curated data, achieving competitive recommendation agreement while maintaining relatively diverse judgments.
\end{enumerate}

\section{Related Work}
\label{sec:related-work}

\textbf{Recursive training and synthetic-data feedback.}
Recursive training on model-generated data can cause
distributional collapse, including the loss of low-probability
modes \citep{shumailov2023curse,alemohammad2024self,dohmatob2024a}.
In language models, synthetic-data exposure can increase
distributional shift and output concentration
\citep{zhu2024synthesize,kovavc2025recursive}.
However, collapse depends on the training regime:
retaining real data alongside synthetic samples can mitigate
or prevent degradation in certain settings
\citep{gerstgrasser2024model,fu2025a}.
We investigate whether analogous effects emerge in scientific
peer review, focusing on recommendation distributions and
semantic diversity under recursive synthetic-review training.

\textbf{LLMs for scholarly peer review.}
LLMs have been applied to review generation, recommendation
prediction, and reviewer assistance
\citep{zhou2024llm,zhuang2025large}.
Datasets such as ReviewMT \citep{tan2024peer} and Gen-Review
\citep{demetrio2025gen} support systematic evaluation,
while ReviewEval \citep{garg2025revieweval} assesses generated
reviews along dimensions including factual accuracy,
analytical depth, and constructiveness.
Prior studies identify limitations such as generic feedback,
insufficiently substantive critiques, and inconsistent
grounding \citep{li2025unveiling,fichtl2026ai,li2026reviewgrounder}.
Meanwhile, LLM-assisted reviewing is increasingly prevalent
\citep{liang2024monitoring}, and carefully designed assistance
can improve review quality \citep{thakkar2025can}.
Unlike prior work evaluating individual AI-generated reviews,
we study how training on such reviews affects subsequent
reviewer models.

\textbf{Test-time activation steering.}
Activation steering modifies model behavior through
interventions on internal representations without updating
model parameters
\citep{li2023inference,zou2023representation,arditi2024refusal}.
Activation Addition \citep{turner2023steering} and
Contrastive Activation Addition \citep{rimsky2024steering}
construct steering directions from contrasting examples.
We adapt contrastive steering to scientific reviewing using
paired official and model-generated reviews of the same
manuscript, investigating whether limited reference
supervision can mitigate judgment homogenization.
\section{Controlled Recursive AI Reviewer Training}\label{sec:recursive-study}

\subsection{Experimental Framework}\label{sec:recursive-framework}

We study one controlled step of recursive AI reviewer training. Starting from a base LLM, we first train a reviewer on earlier official conference reviews. We then use that reviewer to generate synthetic reviews and mix them with official reviews from the following year. This setup isolates how increasing exposure to model-generated judgments affects the next reviewer. We later extend the analysis by one more recursive step in Sec.~\ref{sec:two-step-recursive-training}.

\textbf{Notation.}
Let $M_0$ be the Llama 3.1 8B Instruct model. Let $\mathcal{T}(M,\mathcal{D})$ be supervised fine-tuning of model $M$ on review dataset $\mathcal{D}$. For year $y$, let $\mathcal{D}_{y}^{\mathrm{off}}$ be the official ICLR reviews that year. We use ``official'' rather than ``human'' as these reviews may include unobserved AI assistance.

Let $p\in\mathcal{P}=\{0\%,33\%,66\%,100\%\}$ denote the target fraction of explicitly synthetic reviews in the training corpus.

\textbf{Stage construction.}
Let $\mathcal{D}_{2018:2023}^{\mathrm{off}} = \bigcup_{y=2018}^{2023}\mathcal{D}_{y}^{\mathrm{off}}$. We train the initial reviewer:
\[
M_1 = \mathcal{T}\left(M_0,\mathcal{D}_{2018:2023}^{\mathrm{off}}\right)
\]

We end the initial adaptation stage in 2023 for two
reasons. First, these reviews largely precede the
widespread adoption of ChatGPT-assisted reviewing.
Second, the training-data cutoff of Llama 3.1 8B Instruct
is December 2023 \citep{meta2024llama31}.
Using ICLR 2024 reviews for recursive training therefore
reduces the risk of overlap with the base model's
pretraining data, although such overlap cannot be ruled out.
This temporal separation ensures that the reviews
used to train $M_1$ are distinct from those used to train
its successor models.

For each $p\in\mathcal{P}$, we construct a 2024 training corpus by mixing official reviews with reviews generated by $M_1$ at target synthetic fraction $p$:
\[
\mathcal{D}_{2024}^{p}(M_1)
=
\operatorname{Mix}_{p}\!\left(
\mathcal{D}_{2024}^{\mathrm{off}},
\mathcal{D}_{2024}^{\mathrm{syn}}(M_1)
\right),
\]
where $\operatorname{Mix}_{p}$ denotes corpus-level sampling with target synthetic fraction $p$. The corresponding successor reviewer is
\[
M_2^{p}
=
\mathcal{T}\left(
M_1,
\mathcal{D}_{2024}^{p}(M_1)
\right),
\qquad p\in\mathcal{P}
\]

\noindent\textbf{Controlled recursive exposure.}
All $M_2^{p}$ variants share the same $M_1$ initialization and training configuration. The planned experimental factor is the fraction $p$ of reviews explicitly generated by $M_1$. We use $M_2^{0\%}$ as the official-review only baseline and progressively increase synthetic exposure: $0\% \longrightarrow 33\% \longrightarrow 66\% \longrightarrow 100\%$

Note that, because official reviews may themselves contain unobserved AI assistance, $p$ should not be interpreted as the total fraction of AI-assisted reviews. Rather, it denotes the fraction of training reviews that we explicitly generate using $M_1$. This controlled variation allows us to measure how known increases in synthetic-review exposure affect successor reviewer behavior.

\subsection{Training and Evaluation Setup} \label{sec:recursive-training-llm}

\textbf{Data construction.} We apply the same filtering pipeline at every training stage, removing malformed or follow-up reviews, structurally invalid examples, and examples exceeding the 64k-token context limit.

\noindent\textbf{Training protocol.} We fine-tune all models using LlamaFactory \citep{zheng2024llamafactory} with LoRA \citep{hu2021lora}. $M_1$ is initialized from Meta-Llama-3.1-8B-Instruct and trained on official reviews from 2018--2023 with a learning rate of $2\times10^{-5}$. Each $M_2^{p}$ is initialized from $M_1$ checkpoint and trained for three epochs on $\mathcal{D}_{2024}^{p}(M_1)$ with a learning rate of $2\times10^{-6}$. All $M_2^{p}$ variants use identical optimization and model settings. Detail is in Appx.~\ref{app:training-llm}.

\noindent\textbf{Held-out evaluation set.}
Before constructing the training corpora, we reserve 2,000 papers spanning ICLR 2018--2025 and exclude these papers and their reviews from all training stages. Every trained reviewer generates reviews for this same fixed evaluation set, enabling paired comparison across the $M_2^{p}$ variants.

\section{Scientific-Judgment Collapse under Recursive Training} \label{Sec:Scientific-Judgment Collapse under Recursive Training}

\subsection{Synthetic Exposure Compresses Rating Diversity}
\label{sec:collapse-rating}

\noindent\textbf{Goal.}
We first examine whether increasing synthetic-review exposure changes the distribution of reviewer recommendations. We compare the $M_2^p$ variants trained with increasing proportions of synthetic reviews, using the official-only $M_2^{0\%}$ model as the controlled baseline.

\noindent\textbf{Setup.}
For each model, we extract the overall recommendation from every generated review on the held-out evaluation set and construct the empirical distribution over ratings. We use the entropy of this distribution to measure rating diversity. Higher entropy indicates broader use of the rating scale, while lower entropy indicates stronger concentration. Official review ratings are included only as a contextual reference.

\begin{figure}[t]
\centering
\includegraphics[width=\linewidth]{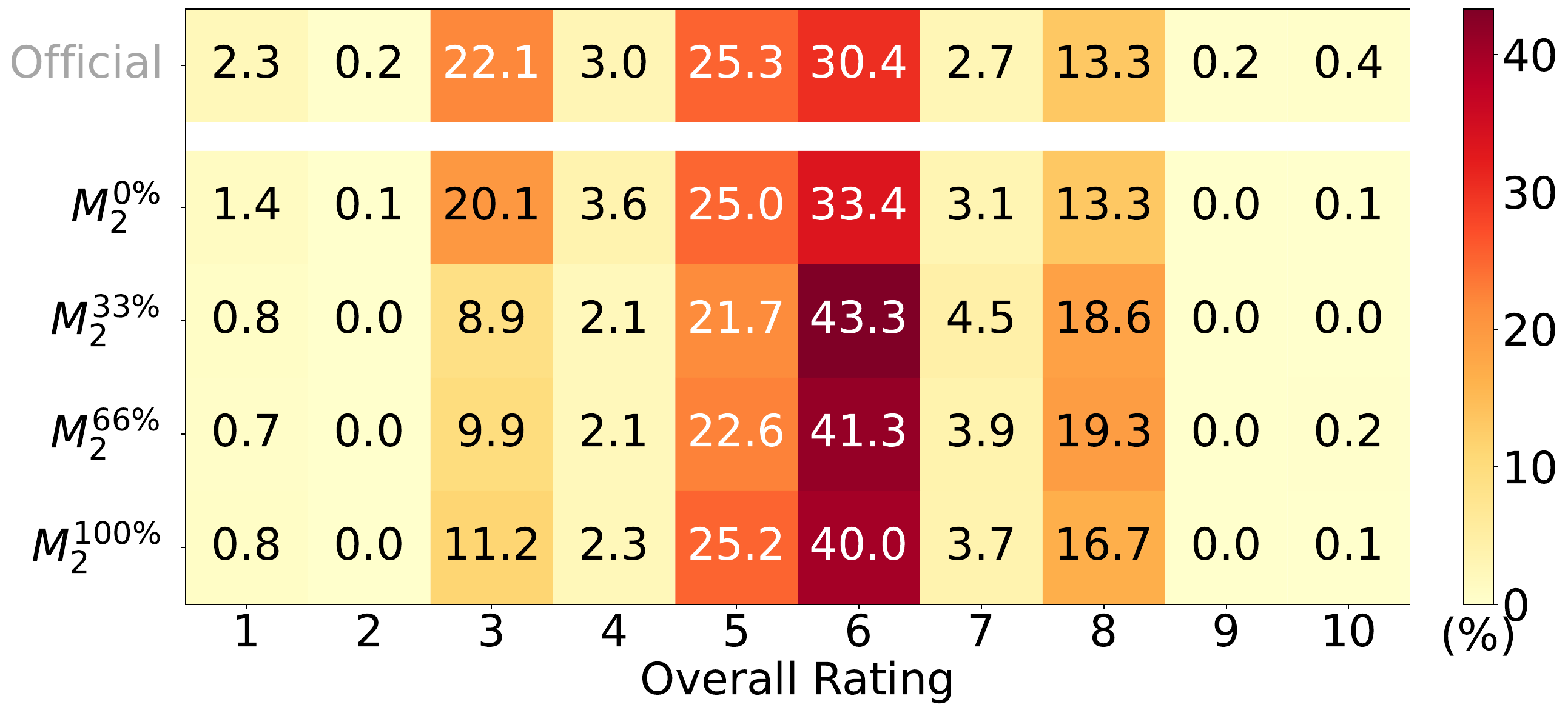}
\caption{
\textbf{Synthetic-review exposure concentrates recommendation ratings.}
Official reviews are included as a contextual reference. Compared with $M_2^{0\%}$, models trained with synthetic reviews assign more recommendations to a narrower range of ratings.
}
\label{fig:judgment-distribution-drift}
\end{figure}

\noindent\textbf{Observation.}
Fig.~\ref{fig:judgment-distribution-drift} shows that recommendation ratings become more concentrated after synthetic reviews are introduced. Compared with $M_2^{0\%}$, $M_2^{33\%}$ assigns more recommendations to rating 6 (from 33.4\% to 43.3\%) and fewer to rating 3 (from 20.1\% to 8.9\%). The 66\% and 100\% variants show similar concentration, with approximately 40\% of recommendations assigned rating 6. Consistently, rating entropy decreases from 2.31 for $M_2^{0\%}$ to 2.14 at 33\% synthetic exposure and remains below the baseline at higher exposure levels. These results show that synthetic-review exposure narrows the distribution of recommendation ratings. Additional analysis in Appx.~\ref{app:recommendation_agreement} shows that lower-rated papers increasingly receive ratings of 5--6.

\begin{figure*}[t]
\centering
\includegraphics[width=\linewidth]{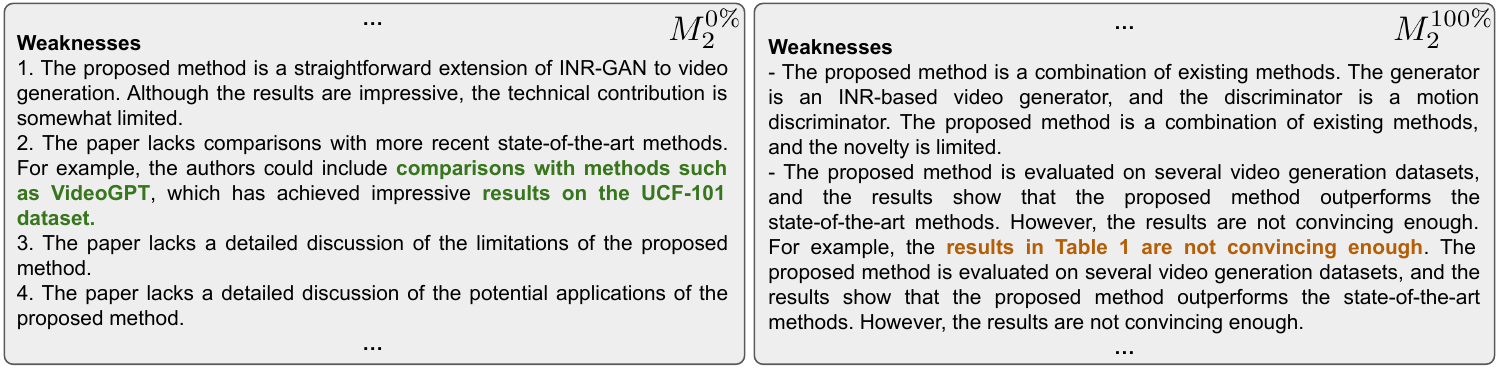}
\caption{\textbf{Qualitative example of review homogenization
under synthetic exposure.}
For the same manuscript, the official-only reviewer
($M_2^{0\%}$) identifies several distinct weaknesses,
whereas the synthetic-trained reviewer ($M_2^{100\%}$)
produces more repetitive and generic critiques.
This example illustrates a qualitative difference
consistent with the measured reduction in semantic diversity.}
\label{fig:qualitative-example}
\end{figure*}

\subsection{Synthetic Exposure Homogenizes Semantic Judgments}
\label{sec:collapse-semantic}

\subsubsection{Same-Paper Semantic Homogenization}

\textbf{Goal.} 
Prior work on diversity evaluation distinguishes semantic
variation from lexical differences in generated text
\citep{tevet2021evaluating}, and motivates evaluating
multiple outputs conditioned on the same input
\citep{chan2024distribution}.
We examine whether increasing synthetic-review exposure
reduces semantic diversity among independently generated
reviews of the same paper.

\textbf{Setup.} For each paper, we generate three independent reviews per model. We encode each review into a semantic embedding and compute the average pairwise cosine distance among generated reviews of the same paper. Details of embedding model and process are in Appx.~\ref{app:semantic-embeddings}. Each paper has review embeddings $\{e_1, \dots, e_n\}$, its same-paper semantic difference is
  \[
  D_{\mathrm{paper}}
  =
  \frac{2}{n(n-1)}
  \sum_{i<j}
  \left(1 - \cos(e_i, e_j)\right).
  \]
Smaller values indicate greater semantic homogeneity. We report the mean $D_{\mathrm{paper}}$ across papers in the held-out evaluation set.

\begin{figure}[t]
\centering
\includegraphics[width=\linewidth]{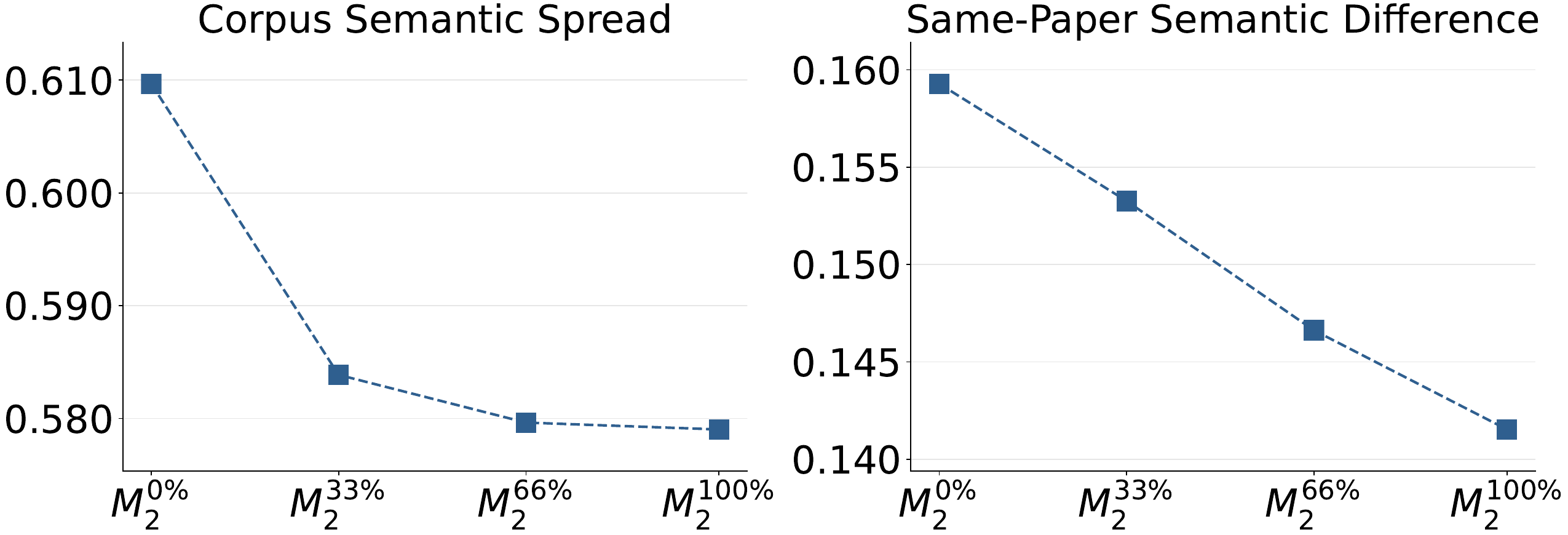}
\caption{
\textbf{Synthetic exposure homogenizes review semantics at both the same-paper and corpus levels.} \textbf{Top:} mean pairwise semantic distance among independently generated reviews of the same submission decreases monotonically with synthetic exposure.
\textbf{Bottom:} semantic spread across paper-level review embeddings also contracts. 
}
\label{fig:semantic-collapse-summary}
\end{figure}

\textbf{Observation.} Fig.~\ref{fig:semantic-collapse-summary}-top shows a monotonic reduction in same-paper semantic diversity as synthetic exposure increases. Within the controlled $M_2$ comparison, mean pairwise distance decreases from $0.159$ for $M_2^{0\%}$ to $0.153$, $0.147$, and $0.142$ at $33\%$, $66\%$, and $100\%$ synthetic exposure, respectively. This corresponds to an approximately $11\%$ reduction from $0\%$ to $100\%$ exposure. This indicates that, as synthetic exposure increases, independently generated reviews of the same submission become progressively more semantically similar.

\subsubsection{Corpus-Level Semantic Contraction}
\label{sec:collapse-corpus-semantic}

\textbf{Goal.} Same-paper analysis measures conditional variation when the submission is held fixed. We next ask whether the broader semantic space occupied by reviews across different papers also contracts under synthetic exposure.

\textbf{Metric.} For each review, we encode the full review into a semantic embedding and aggregate multiple reviews of the same paper into a single paper-level embedding (See Appx.\ref{app:semantic-embeddings}). We then measure how different these paper-level embeddings are in semantic space. Let $\{e_1,\ldots,e_N\}$ be the paper-level embeddings, the corpus centroid is
\[
\bar{e}=\frac{1}{N}\sum_{i=1}^{N}e_i.
\]
We define corpus semantic spread as
\[
\mathrm{Spread}
=
\frac{1}{N}\sum_{i=1}^{N}\|e_i-\bar{e}\|_2^2.
\]
Smaller values indicate greater concentration.

\textbf{Observation.} Fig.~\ref{fig:semantic-collapse-summary}-bottom shows a  consistent contraction at the corpus level. Semantic spread decreases from $0.609$ for $M_2^{0\%}$ to $0.584$, $0.580$, and $0.579$ at $33\%$, $66\%$, and $100\%$ synthetic exposure, respectively. This corresponds to an approximately $5\%$ reduction from $M_2^{0\%}$ to $M_2^{100\%}$. Most of the contraction occurs when synthetic supervision is first introduced, with smaller additional decreases at higher exposure levels.

Together, the same-paper and corpus-level analyses show that synthetic exposure reduces semantic diversity both: among reviews of the same submission, and globally across the evaluation corpus.

\begin{figure*}[t]
\centering
\includegraphics[width=.96\linewidth]{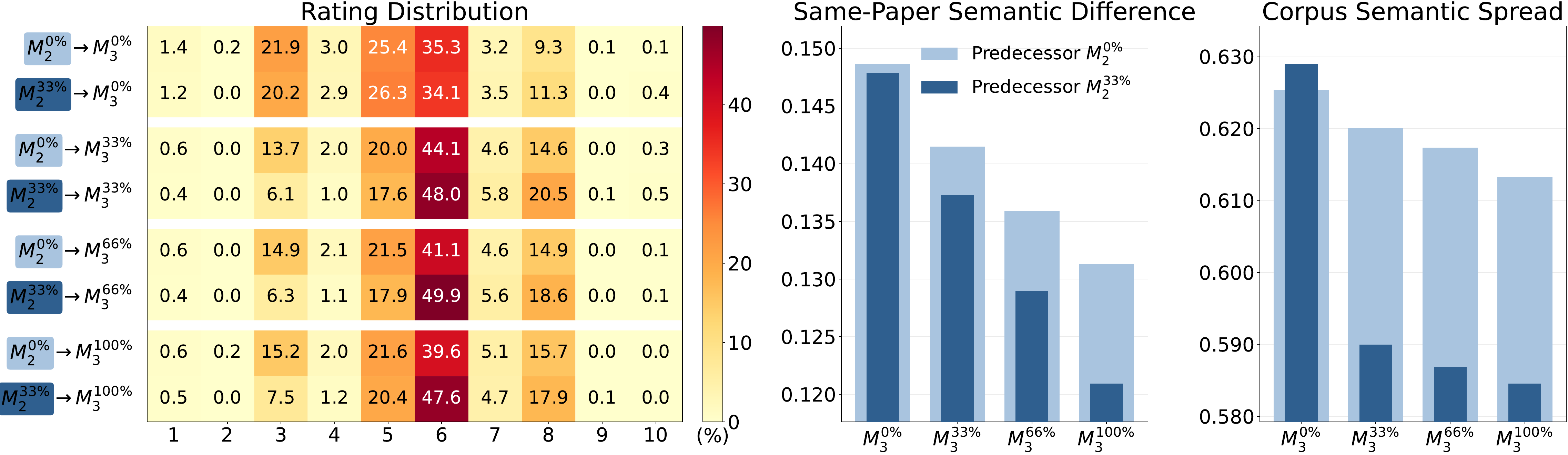}
\vspace{-0.1cm}
\caption{
\textbf{Prior synthetic exposure amplifies judgment homogenization.} We compare reviewers trained with and without prior synthetic exposure ($M_2^{33\%}$ vs. $M_2^{0\%}$), while keeping current-stage exposure $p$ the same. \textbf{Left:} recommendation-rating distributions. \textbf{Middle:} semantic diversity among reviews of the same paper. \textbf{Right:} semantic diversity across papers. At $p=0\%$, both groups show similar semantic diversity. When synthetic supervision is reintroduced, reviewers with prior synthetic exposure produce more homogeneous judgments.
}
\label{fig:m3-prior-exposure}
\end{figure*}

\subsection{Prior Recursive Exposure Amplifies Judgment Homogenization}
\label{sec:two-step-recursive-training}

\textbf{Goal.}
Results in Sec.~\ref{sec:collapse-semantic}/~\ref{sec:collapse-rating} show that synthetic-review exposure homogenizes scientific judgments after one-step recursive training. We next study two-step recursive training and ask if prior synthetic exposure changes a reviewer's response to synthetic supervision at the following stage. Specifically, at matched current-stage synthetic exposure, we compare successors whose predecessors were trained with versus without synthetic reviews.

\textbf{Setup.}
We construct two $M_3$ lineages. In the no-prior-exposure lineage, the predecessor is $M_2^{0\%}$, which received no explicit synthetic supervision during the preceding stage. In the prior-exposure lineage, the predecessor is $M_2^{33\%}$, which was previously trained with synthetic reviews. From each predecessor, we train successor reviewers $M_3^{p}$ at matched current-stage synthetic fractions $p\in\{0\%,33\%,66\%,100\%\}$.

We denote the two transitions as $M_2^{0\%}\!\rightarrow M_3^{p}$ and $M_2^{33\%}\!\rightarrow M_3^{p}$, where the $M_2$ superscript indicates prior synthetic exposure and $p$ indicates current-stage exposure. We compare them using the same same-paper semantic distance and corpus-level semantic spread defined in Sec.~\ref{sec:collapse-semantic}. Because each predecessor also generates the synthetic reviews used to train its successor, this comparison captures recursive training history as a whole rather than an initialization effect alone.

\textbf{Observation.}
Fig.~\ref{fig:m3-prior-exposure} shows that the effect of current synthetic training depends on prior synthetic exposure. At $0\%$ current exposure, the two lineages have nearly identical same-paper semantic distance, while corpus-level spread is slightly higher for the prior-exposure lineage. Prior exposure alone does not make the successor consistently more homogeneous after another stage of official-review training. 

However, once synthetic supervision is reintroduced, successors of $M_2^{33\%}$ become consistently more homogeneous than successors of $M_2^{0\%}$. At matched current-stage exposures of $33\%$, $66\%$, and $100\%$, same-paper semantic distance is approximately $3.0\%$, $5.1\%$, and $7.9\%$ lower, respectively, for the $M_2^{33\%}\!\rightarrow M_3^{p}$ transition. Corpus-level semantic spread shows a similar pattern. Recommendation ratings become more concentrated, particularly around rating 6. Together, these suggest that prior synthetic-review exposure increases a successor reviewer's susceptibility to judgment homogenization under subsequent synthetic training.
\section{Inference-Time Recovery from Limited Reference Supervision}
\label{sec:inference-time-recovery}

\begin{table*}[t]
\centering
\caption{Inference-time recovery using activation steering. We compare the official-only baseline ($M_2^{0\%}$), the highest synthetic-exposure model ($M_2^{100\%}$), and the same model with our steering vector. Higher entropy and semantic diversity indicate more diverse judgments, while higher exact match and lower MAD indicate better agreement with official recommendations.}
\label{tab:m2-steering-mitigation}
\resizebox{1.0\columnwidth}{!}{
\begin{tabular}{lccccc}
\toprule
\textbf{System}
& \textbf{Exact Match} $\uparrow$
& \textbf{MAD} $\downarrow$
& \textbf{Rating Entropy} $\uparrow$
& \textbf{Corpus Spread} $\uparrow$
& \textbf{Same-Paper Diversity} $\uparrow$ \\
\midrule
$M_2^{0\%}$
& $74.55 \pm 0.45$
& $1.146 \pm 0.019$
& $2.311 \pm 0.017$
& $0.610 \pm 0.001$
& $0.159$ \\

$M_2^{100\%}$
& $72.53 \pm 0.59$
& $1.152 \pm 0.024$
& $2.184 \pm 0.007$
& $0.579 \pm 0.001$
& $0.142$ \\

$M_2^{100\%}$ + Steering
& $\mathbf{73.42 \pm 0.35}$
& $\mathbf{1.128 \pm 0.013}$
& $\mathbf{2.201 \pm 0.010}$
& $\mathbf{0.585 \pm 0.005}$
& $0.132$ \\
\bottomrule
\end{tabular}
}
\end{table*}

We investigate whether scientific-judgment collapse can be partially reversed at inference time using a limited set of reference reviews. Rather than retraining the collapsed reviewer, we use contrastive activation steering to guide its judgments toward those of the reference reviews while keeping model parameters fixed.


\textbf{Contrastive Activation Steering.} Let $M_{\mathrm{col}}$ be the collapsed reviewer after recursive training. We construct a calibration set $\mathcal{S}_{K} = \{(x_i,r_i^{\star})\}_{i=1}^{K}$, where $x_i$ is a paper and $r_i^{\star}$ is a reference review of that paper. For each $x_i$, $M_{\mathrm{col}}$ generates a corresponding review $\hat{r}_i$. We use pairs of $(\hat{r}_i,r_i^{\star})$ to construct the contrastive steering vector. Particularly, for layer $\ell$, let $\phi_{\ell}(r)$ denote the final hidden state obtained by processing review $r$ with $M_{\mathrm{col}}$. We estimate a reference-oriented steering direction by averaging the paired representation differences:
\[
v_{\mathrm{steer}}^{(\ell)}
=
\frac{1}{K}
\sum_{i=1}^{K}
\left[
\phi_{\ell}(r_i^{\star})
-
\phi_{\ell}(\hat{r}_i)
\right].
\]
This direction captures the average representation difference between reference reviews and reviews by the collapsed model for the same papers. At inference time, for selected layers $\ell\in\mathcal{L}$, we add the steering direction to the residual-stream activation at the current generation position:
\[
\widetilde{h}_{\ell,t}
=
h_{\ell,t}
+
\alpha v_{\mathrm{steer}}^{(\ell)},
\]
where $\alpha$ controls the steering strength. The intervention modifies activations during generation but leaves the model parameters unchanged. The calibration papers used to estimate $v_{\mathrm{steer}}^{(\ell)}$ are disjoint from all evaluation papers.


\textbf{Experimental Setup.} We use $M_2^{100\%}$ as the collapsed reviewer $M_{\mathrm{col}}$, corresponding to the highest synthetic-exposure condition in our controlled experiment. We randomly sample 500 papers (excluded from held-out evaluation set) as $\mathcal{S}_{K}$. We set $\alpha=0.25$ and steer only last layer. The details of steering hyperparameter selection is provided in Appx.~\ref{app:steering-selection}.

\textbf{Observation.}
Tab.~\ref{tab:m2-steering-mitigation} shows that activation steering partially recovers the effects of recursive training. Compared with $M_2^{100\%}$, steering improves exact recommendation match from 72.53\% to 73.42\%, increases rating entropy from 2.184 to 2.201, and increases semantic diversity across papers from 0.579 to 0.585. However, semantic diversity among independently generated reviews of the same paper decreases from 0.142 to 0.132. These suggest that steering can partially reverse judgment homogenization, but not across all measures.

\section{TrustReviewer: A Practical System from Our Study}
\label{sec:trustreviewer}

We explore steering approach beyond controlled experiments and develop TrustReviewer. Particularly, we train the model on a large-scale curated dataset and apply activation steering at inference time.

\subsection{System Design}
\label{sec:trustreviewer-design}

\textbf{Curated official-review training corpus.}
We collect officially released ICLR reviews from 2018 to 2025: $\mathcal{D}_{\mathrm{raw}} = \bigcup_{y=2018}^{2025}\mathcal{D}_{y}^{\mathrm{off}}$. As throughout this work, we use ``official'' rather than ``human'' because recent reviews may contain unobserved AI assistance. We apply the filtering and curation pipeline (See Appx.~\ref{app:data-filtering}), removing malformed, duplicated, excessively short, repetitive, and follow-up reviews. The resulting corpus contains 112,743 paper-review pairs and approximately 1.9 billion tokens. All held-out evaluation papers are excluded from corpus construction and training.

We initialize the core reviewer $M_{\mathrm{TR}}$ from Meta-Llama-3.1-8B-Instruct \citep{meta2024llama31} and fine-tune it directly on the curated corpus using the training protocol in Sec.~\ref{sec:recursive-training-llm}. Unlike the controlled recursive-training experiments in Sec.~\ref{sec:recursive-study}, TrustReviewer does not undergo a recursive synthetic-review training stage. We treat this as a practical design choice motivated by our analysis rather than causal evidence that curation alone prevents judgment collapse.

\textbf{Activation steering.}
At inference time, we apply the paired activation-steering procedure from Sec.~\ref{sec:inference-time-recovery}. We use 5,000 randomly sampled examples from training to estimate a reference-oriented activation direction, which is then applied during generation while leaving model parameters unchanged. Steering hyperparameter selection is in Appx.~\ref{app:steering-selection}.






\begin{table*}[t]
\centering
\caption{
\textbf{Held-out reviewer evaluation.}
Exact match and MAD measure agreement with official recommendations. Entropy measures rating-scale usage, while corpus spread and same-paper distance measure semantic diversity. 
Best results are shown in bold and second-best results are underlined.
}
\label{tab:trustreviewer-results}
\resizebox{\textwidth}{!}{%
\begin{tabular}{lccccc}
\toprule
& \multicolumn{3}{c}{Judgment Alignment}
& \multicolumn{2}{c}{Semantic Diversity} \\
\cmidrule(lr){2-4}
\cmidrule(lr){5-6}
Model
& \shortstack{Exact Match\\(\%) $\uparrow$}
& MAD $\downarrow$
& Entropy $\uparrow$
& \shortstack{Corpus\\Spread $\uparrow$}
& \shortstack{Same-Paper\\Distance $\uparrow$} \\
\midrule
Meta-Llama-3.1-8B-Instruct
& $33.93 \pm 0.54$
& $2.679 \pm 0.010$
& $1.53 \pm 0.03$
& $0.5718 \pm 0.0007$
& $0.0751$ \\

Qwen3.6-35B-A3B
& $61.85 \pm 0.33$
& $1.335 \pm 0.012$
& $1.94 \pm 0.02$
& $0.5570 \pm 0.0016$
& $0.0482$ \\

OpenReviewer
& $73.10 \pm 1.36$
& $1.113 \pm 0.022$
& $2.10 \pm 0.05$
& $\mathbf{0.6367 \pm 0.0147}$
& $\mathbf{0.1695}$ \\

TrustReviewer w/o Steering
& $\underline{73.85 \pm 0.15}$
& $\mathbf{1.077 \pm 0.005}$
& $\underline{2.13 \pm 0.03}$
& $0.6189 \pm 0.0004$
& $\underline{0.1520}$ \\

TrustReviewer
& $\mathbf{75.40 \pm 0.46}$
& $\underline{1.079 \pm 0.005}$
& $\mathbf{2.18 \pm 0.01}$
& $\underline{0.6301 \pm 0.0030}$
& $0.1481$ \\
\bottomrule
\end{tabular}%
}
\end{table*}

\subsection{Evaluation}
\label{sec:trustreviewer-evaluation}

\textbf{Baseline.} We evaluate TrustReviewer on the same held-out paper set used throughout our analysis. We compare against Meta-Llama-3.1-8B-Instruct \citep{meta2024llama31}, the initialization used for TrustReviewer; Qwen3.6-35B-A3B \citep{qwen36_35b_a3b}; and OpenReviewer \citep{idahl2025openreviewer}, an open-source reviewer specialized for scientific evaluation using the same LLM backbone with TrustReviewer. We additionally compare TrustReviewer with and
without inference-time steering.

\textbf{Metrics.} We measure recommendation agreement using exact match and mean absolute distance (MAD) from the average official rating metrics following \citep{idahl2025openreviewer}. We characterize judgment diversity using rating entropy, same-paper semantic distance, and corpus-level semantic spread. Results are aggregated over three independent generation runs. The complete generation and parsing details are provided in Appx.~\ref{app:generation-and-parsing}.

\subsection{Results}
\label{sec:trustreviewer-results}

\textbf{TrustReviewer achieves strong recommendation agreement.} Tab.~\ref{tab:trustreviewer-results} shows that TrustReviewer without steering achieves an exact-match rate of $73.85\%$, compared with $73.10\%$ for OpenReviewer, $61.85\%$ for Qwen3.6-35B-A3B, and $33.93\%$ for Meta-Llama-3.1-8B-Instruct. It also obtains the lowest
MAD.

\textbf{TrustReviewer maintains relatively diverse judgments.} Without steering, TrustReviewer reaches a rating entropy of $2.13$, exceeding Llama ($1.53$), Qwen ($1.94$), and OpenReviewer ($2.10$). For context, official reviews have rating entropy 2.38. TrustReviewer maintains substantially greater semantic diversity than the general-purpose Llama and Qwen baselines and remains competitive with the specialized OpenReviewer.

\textbf{Inference-time steering provides additional gains.}
Applying paired activation steering increases exact recommendation match from $73.85\%$ to $75.40\%$ and rating entropy from $2.13$ to $2.18$. MAD remains essentially unchanged. Steering also increases
corpus-level semantic spread, although same-paper semantic distance
decreases slightly. Thus, steering improves recommendation agreement
and aggregate rating diversity while largely preserving the strong
diversity of the core reviewer.

Overall, TrustReviewer provides a practical system informed by our study:
it trains directly on curated official-review supervision rather than
recursive synthetic reviews and can incorporate limited reference
supervision through a parameter-fixed inference-time intervention.
The resulting reviewer achieves strong recommendation agreement while
maintaining relatively diverse scientific judgments.

\section{Conclusion}

We study how recursive AI-reviewer training affects scientific judgment. In controlled experiments, synthetic-review exposure concentrates recommendation ratings and reduces same-paper and corpus-level semantic diversity. Extending the training chain shows that prior synthetic exposure increases susceptibility to homogenization when synthetic supervision is reintroduced. We refer to this failure mode as \emph{scientific-judgment collapse}. We further explore whether limited reference supervision can partially recover judgment diversity at inference time without updating model parameters. Finally, TrustReviewer instantiates these insights in a practical system trained once on curated official reviews, achieving strong recommendation agreement while maintaining relatively diverse judgments. Our findings do not imply that recursive collapse is inevitable or that greater diversity necessarily means better review quality.

\textbf{Limitations.} Our experiments cover two recursive training steps, one base-model family, and one scientific-review domain, limiting generalizability. Following \cite{idahl2025openreviewer}, our metrics are proxies for scientific judgment rather than review quality. Official ratings may contain disagreement or AI assistance, while embedding-based diversity may capture stylistic rather than substantive differences. Future work should assess whether recursive training suppresses valid critiques or reduces complementary information through human evaluation.

\section{Ethical Considerations}

LLM-based scientific reviewing may reinforce existing biases, contribute to scientific-judgment collapse. Although TrustReviewer aims to improve recommendation agreement and judgment diversity, these improvements do not guarantee review quality or fairness. TrustReviewer is intended to assist rather than replace human reviewers.

\section*{Acknowledgments}

Ho, Liu, and Huang are supported by DARPA HR001124S0029-AIQ-FP-019 and the National Science Foundation TRAILS Institute (2229885). Private support was provided by Open Philanthropy and Apple. The authors acknowledge the National Artificial Intelligence Research Resource (NAIRR) Pilot for supporting this research.

\bibliography{ref}

\appendix

\onecolumn
\section*{Appendix}
\section{AI Use Disclosure}

Generative AI was used as part of the research methodology to generate synthetic reviews and model outputs, as described in the paper. Separately, generative AI assisted with language editing, manuscript organization, title development, presentation of existing results, and editing experimental code. It did not derive results or create citations. The authors verified all revisions and take responsibility for the text, claims, and artifacts.

\section{Dataset and Model Licensing.} We use publicly available ICLR reviews collected from OpenReview under CC BY 4.0 and Llama 3.1 8B under the Llama 3.1 Community License. We release our curated review corpus and fine-tuned model in accordance with the applicable attribution and redistribution requirements.

\textbf{Data Privacy and Content.} Our dataset consists of publicly available ICLR reviews from OpenReview. We exclude unnecessary personally identifiable information from the released training corpus. Although scientific reviews are not expected to contain offensive content, we acknowledge that publicly submitted reviews may contain inappropriate language.

\section{Recommendation Agreement under Synthetic Exposure}
\label{app:recommendation_agreement}

\begin{figure}[ht]
    \centering
    \includegraphics[width=.75\textwidth]{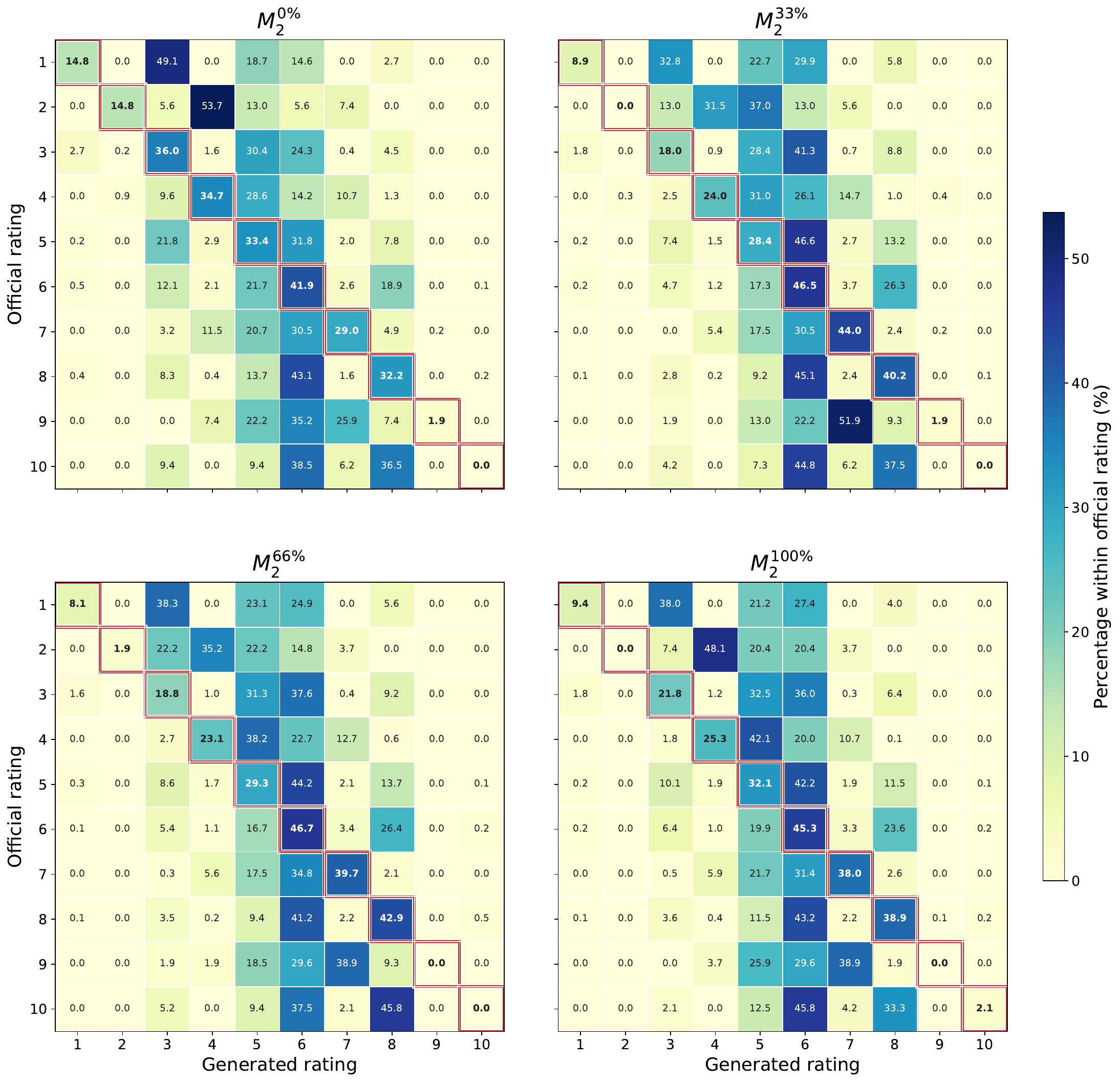}
    \caption{
        \textbf{Recommendation ratings become more concentrated under synthetic-review exposure.} Row-normalized confusion matrices compare generated and official ratings across the four $M_2$ variants. Each row shows how the model rates papers with a given official rating. As synthetic exposure increases, papers with lower official ratings are increasingly assigned ratings of 5--6.
    }
    \label{fig:m2_rating_confusion}
\end{figure}

To further examine recommendation concentration, Fig.~\ref{fig:m2_rating_confusion} compares generated and official ratings across synthetic-review exposure levels. We observe that increasing synthetic exposure shifts recommendations toward ratings 5--6, particularly for lower-rated papers. For example, among assignments with an official rating of 3, the proportion receiving ratings 5--6 increases from 54.7\% to 68.5\%, while exact matches decrease from 36.0\% to 21.8\%.

\section{Detailed Data Curation}\label{app:data-filtering}

\paragraph{Paper inputs.} We collect ICLR papers and their reviews from 2018 to 2025 via the OpenReview API. We then use Marker\footnote{\url{https://github.com/datalab-to/marker}} to convert the paper PDFs to Markdown. We retain only the main paper text and discard appendix content. 

\paragraph{System prompt and review forms.} Following the guideline template in \citet{idahl2025openreviewer}, the system prompt assigns the model the role of an AI-conference reviewer and provides general instructions for constructive, evidence-based evaluation. It asks the model to assess the paper's objective, strengths, weaknesses, correctness, novelty, significance, and supporting evidence. 

The requested review fields are not fixed across all examples. We insert the ICLR form corresponding to the submission year into the system prompt. For example, earlier forms request a free-form review and overall rating, while later forms additionally specify structured fields such as correctness, technical and empirical novelty, soundness, presentation, contribution, strengths, weaknesses, questions, and ethics concerns. The rating scale and permitted rating values are likewise taken from the corresponding annual form. Thus, the model is trained to follow the review structure that was used for the original supervision.

\paragraph{User prompt.} The user prompt is deliberately minimal:
  \begin{quote}
  \small
  Please review the following paper according to the ICLR reviewing guidelines.

  \texttt{\{paper text\}}
  \end{quote}

\paragraph{Data filtering.} We apply the same two-stage filtering pipeline to every review dataset before training.

\textit{Quality and structural filtering.} In the first stage, we retain a review only if it contains a parseable rating and follows the year-specific review structure derived from the corresponding conference form. We exclude reviews that satisfy any of the following conditions:
  \begin{itemize}
      \item contain follow-up response markers such as ``follow-up comment'' or ``revision'';
      \item contain fewer than 600 characters;
      \item contain at least 8 duplicated lines;
      \item contain at least 4 repeated 30-grams, indicating degenerate repetition; or
      \item fail the required year-specific review structure.
  \end{itemize}

\textit{Token-length filtering.} In the second stage, we format each paper--review pair as the complete training chat sequence and retain it only if its length is strictly below 64{,}000 tokens. Applying these criteria consistently yields a more structurally uniform corpus while reducing malformed, noisy, repetitive, and overlength supervision.

\section{LLM Training protocol} \label{app:training-llm} 

We use supervised fine-tuning in LlamaFactory \citep{zheng2024llamafactory} with LoRA \citep{hu2021lora}, using rank 64, a scaling factor of 128, dropout of 0.05, and adaptation of all target modules. Shared settings for $M_1$ and all $M_2^{p}$ variants include a maximum context length of 64{,}000 tokens, a per-device batch size of 1, 16 gradient-accumulation steps, a warmup ratio of 0.05, three epochs, bfloat16 precision, and FlashAttention-2. We initialize $M_1$ from $M_0$ (Meta-Llama-3.1-8B-Instruct) and fine-tune it on filtered official reviews from 2018--2023 with a learning rate of $2\times10^{-5}$ and cosine decay. Each $M_2^{p}$ is initialized from $M_1$ and fine-tuned on $\mathcal{D}_{2024}^{p}(M_1)$ with a lower learning rate of $2\times10^{-6}$ for $p\in\mathcal{P}$. All $M_2^{p}$ variants share the same training settings, isolating the effect of synthetic-review percentage $p$ on reviewer behavior. For $M_3^{p}$ variants, we use the same training setting as $M_2^{p}$.

All experiments were conducted on a Slurm computing cluster using NVIDIA H200 and RTX A6000 GPUs. Model training consumed approximately 549 GPU-hours on H200 GPUs. Review generation, steering, and evaluation were performed on RTX A6000 GPUs. 

\section{Held-out evaluation set.} \label{app:held-out-eval-set}

We reserve a held-out evaluation set of 2,000 papers sampled across years: 57, 88, 136, 161, 161, 235, 449, and 713 papers from 2018, 2019, 2020, 2021, 2022, 2023, 2024, and 2025, respectively. These papers are excluded from training. Each trained reviewer model generates reviews for this fixed set, and all analysis is performed on those generated reviews. 

\section{Semantic Embedding Details}
  \label{app:semantic-embeddings}

For all semantic embedding analyses, we use the \texttt{jinaai/jina-embeddings-v3} \citep{sturua2024jina} with its symmetric \texttt{text-matching} encoding task. This model supports sequences of up to 8,192 tokens. We set the maximum embedding length to 8,192 tokens. All 37,681 held-out official and generated reviews used in our semantic analyses fit within this limit (maximum 5,940 tokens). Therefore, no review is truncated.

We obtain one L2-normalized embedding per review using the model's task-aware encode interface. Cosine similarity between two review embeddings is computed as their dot product.

For corpus-level analyses, we embed each review independently. When multiple official reviews are available for the same paper, we average their normalized review embeddings and L2-normalize the resulting paper-level vector before computing corpus-level semantic distances. The same-paper analysis instead computes pairwise cosine distances directly between normalized embeddings of individual reviews generated for the same paper.

\section{Steering Hyperparameter Selection}
\label{app:steering-selection}

We select steering hyperparameters separately for TrustReviewer and the controlled $M_2^{100\%}$ mitigation experiment because their steering vectors are constructed from different models and calibration corpora. We select the steering layer set $\mathcal{L}$ and strength $\alpha$ using 100 validation paper--review pairs sampled from $\mathcal{D}_{\mathrm{curated}}$. These pairs are separate from the calibration pairs used to construct the steering vectors, and neither set contains papers from the held-out evaluation set. The selection criterion is exact recommendation match against the corresponding official review ratings.

\paragraph{TrustReviewer.}
We construct the TrustReviewer steering vector from 5,000 non-evaluation papers. The positive member of each pair is an official review. The negative member is a review generated by the unsteered TrustReviewer model for the same paper.

We first fix $\alpha=0.10$ and compare steering at all decoder layers, progressively later layer ranges (16--31, 20--31, 24--31, and 28--31), and the final layer alone (layer 31). This search selects $\mathcal{L}^{*}_{\mathrm{TR}}=\{31\}$. We then evaluate $\alpha\in\{0.05,0.10,0.15,0.20,0.25,0.30,0.40,0.50\}$ at layer 31. This selects $\alpha^{*}_{\mathrm{TR}}=0.15$. TrustReviewer therefore uses $(\mathcal{L}^{*}_{\mathrm{TR}},\alpha^{*}_{\mathrm{TR}}) =(\{31\},0.15)$.

\paragraph{Controlled $M_2^{100\%}$ mitigation.} We use 500 non-evaluation papers, pairing one official review with one review generated by $M_2^{100\%}$ for the same paper.

We repeat the same layer and strength searches. This search selects $(\mathcal{L}^{*}_{M_2},\alpha^{*}_{M_2})=(\{31\},0.25)$.


\section{Review Generation and Recommendation Parsing}
\label{app:generation-and-parsing}

\paragraph{Generation settings.}
For locally generated reviews, we sample with temperature 0.6, \texttt{top\_p} 0.95, a repetition penalty of 1.0, and a maximum of 4,096 new tokens. The scripts do not explicitly set \texttt{top\_k}. We generate three samples per paper in three separate runs. The generation scripts do not use fixed decoding seeds, which also affects paper order and the selection of alternative user-prompt phrasing. The reported variation therefore reflects three independent generation runs rather than deterministic, seed-labeled replications. Fixed seeds used by downstream analysis scripts for sampling, grouping, or bootstrapping are separate from the generation process.

\paragraph{Prompt and paper text.}
The review prompt combines a general system prompt, the year-specific ICLR review form, and a user prompt containing the paper text. Papers are converted to Markdown and truncated after the references section. Local baselines receive the same held-out paper identifiers, Markdown paper text, prompt structure, and default decoding settings unless a model-specific wrapper overrides a setting.

\paragraph{Recommendation parsing and invalid outputs.}
The strict parser locates a line containing the structured \texttt{\#\# Rating} heading, with an optional colon, and reads the beginning of the immediately following line as either an integer rating or \texttt{NA}. During generation, a malformed response is retried up to 10 times, and a review is written only when the strict parser finds a valid numerical rating. If all attempts fail, the paper is recorded as failed for that generation run.

For downstream analysis, the optional relaxed parser additionally searches the rating block for expressions such as \texttt{Rating: 6}, \texttt{Score: 6}, \texttt{6 out of 10}, or \texttt{overall score of 6}. A missing, unparseable, or \texttt{NA} rating yields no numerical recommendation and is never mapped to a default value. Such cases are recorded as an invalid generated rating, missing official reviews, or no valid official ratings, as applicable, and are excluded from score-based metrics.

\begin{figure*}[t]
\centering
\includegraphics[width=\linewidth]{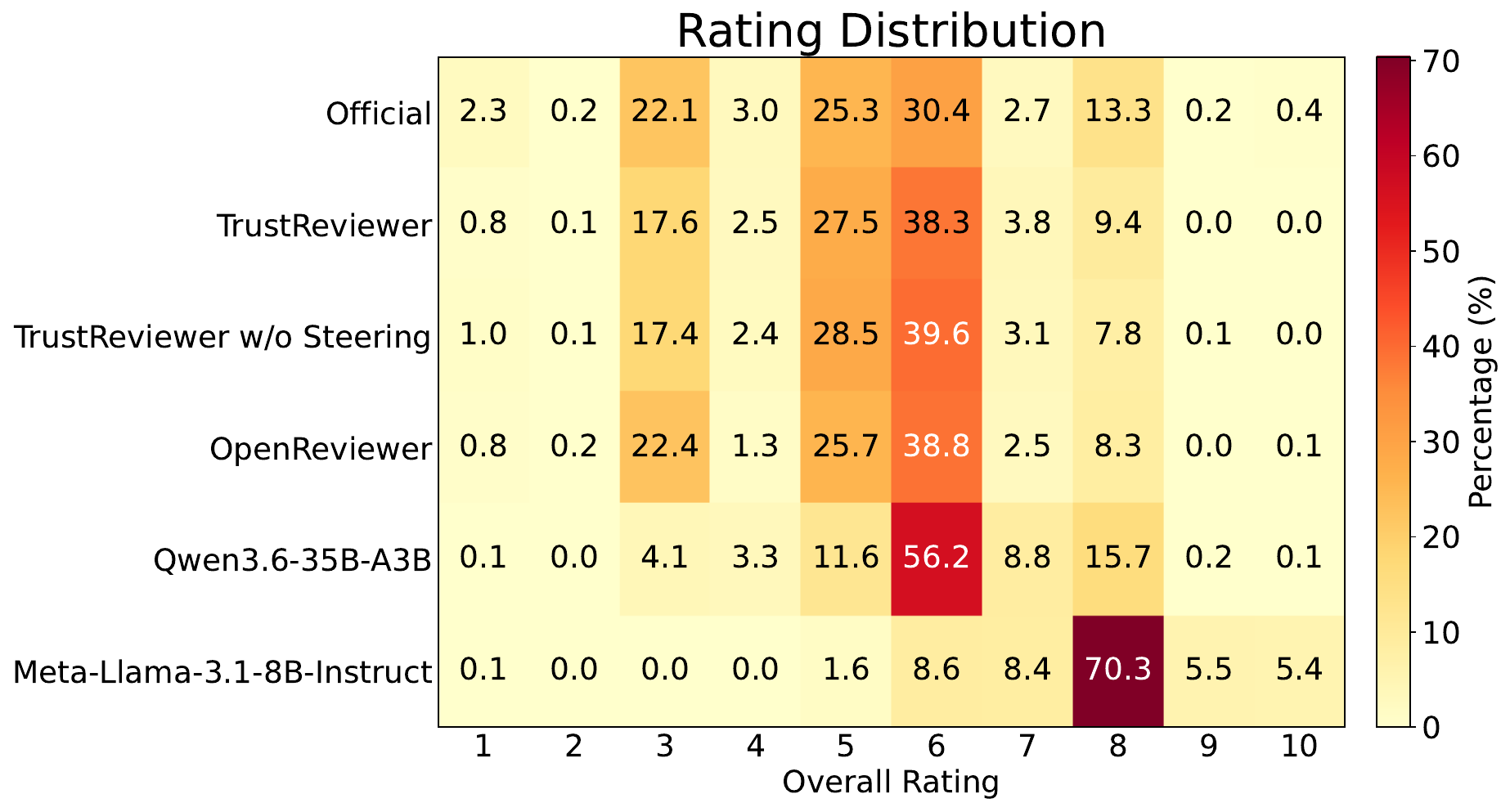}
\caption{\textbf{Rating distributions across reviewers.} The heatmap reports the percentage of reviews assigned to each overall rating for TrustReviewer with and without steering and the baseline models. Official reviews in the plot provide a reference distribution.}
\label{fig:reviewer-rating-comparison}
\end{figure*}

\section{Recommendation Agreement Metrics}
\label{app:recommendation-metrics}

Following \citet{idahl2025openreviewer}, we measure agreement between generated recommendations and official reviewer ratings using two metrics. For each generation run, the valid evaluation set contains papers with both a parsed numerical recommendation and at least one valid official rating. Let $N$ be the size of this set and $m_i$ the number of valid official ratings available for paper $i$.

\begin{itemize}[leftmargin=1.2em, labelsep=0.4em]
    \item Exact match. Let $\hat{s}_i$ denote the generated recommendation for paper $i$, and let $\mathcal{S}_i=\{s_{i1},\ldots,s_{im_i}\}$ denote its official reviewer ratings. We measure exact recommendation matching as
        \[
        \mathrm{Match}
        =
        \frac{1}{N}
        \sum_{i=1}^{N}
        \mathbf{1}\!\left[\hat{s}_i\in\mathcal{S}_i\right].
        \]
    \item To capture the magnitude of disagreement, we also report the mean absolute distance from the average official rating:
        \[
        \mathrm{MAD}
        =
        \frac{1}{N}
        \sum_{i=1}^{N}
        \left|
        \hat{s}_i-\frac{1}{m_i}\sum_{j=1}^{m_i}s_{ij}
        \right|.
        \]
\end{itemize}

Higher matching and lower MAD indicate closer agreement with the official recommendations. These metrics assess recommendation alignment rather than the factual correctness or completeness of the review text. We compute each metric separately for the three generation runs and report the mean and standard deviation across runs.

\section{Reviewer Rating Distributions}
\label{app:reviewer-rating-distributions}

See Fig.~\ref{fig:reviewer-rating-comparison} for rating distributions across LLM-based reviewer with official reviews as reference.





\section{More Limitations Discussion}
\label{app:limitations}

Our controlled experiments examine up to two recursive training steps using one base-model family and one scientific-review domain. The findings may not generalize to longer training chains, other model families and scales, or different scientific fields and review settings.

Our metrics are proxies for scientific judgment rather than direct measures of review quality. Recommendation agreement relies on official ICLR ratings, which may contain disagreement, errors, or unobserved AI assistance. Embedding-based diversity measures may capture stylistic rather than substantive differences in critiques. Importantly, greater diversity does not necessarily imply better reviews, as diverse judgments may be incorrect and homogeneous judgments may be accurate \citep{alihosseini2019jointly,tevet2021evaluating}.

Future work should investigate whether recursive training suppresses valid minority critiques or reduces complementary information across reviewers. Human evaluation of critique correctness, specificity, and coverage would help establish the practical consequences of scientific-judgment homogenization.

\end{document}